\documentclass[11pt]{article}

\usepackage{amsmath,amssymb,natbib,graphicx,url,algorithm2e, hyperref}
\usepackage[T1]{fontenc}
\usepackage[a4paper]{geometry}
\usepackage[font={small},labelfont=bf,format=hang]{caption}
\usepackage{hyperref}
\usepackage{cleveref}
\usepackage{array}
\usepackage{enumitem}
\usepackage{textcomp}
\usepackage{chngcntr}

\hypersetup{
    colorlinks=true,
    linkcolor=black,
    citecolor=black,
    urlcolor=blue,
}

\newcommand{\sectionref}[1]{Section~\ref{#1}}
\newcommand{\figureref}[1]{Figure~\ref{#1}}
\newcommand{\tableref}[1]{Table~\ref{#1}}
\newcommand{\appendixref}[1]{Appendix~\ref{#1}}

\title{\vspace{-40pt}\textbf{Flatland Competition 2020: MAPF and MARL for Efficient Train Coordination on a Grid World}\vspace{20pt}}

\author{
    Florian Laurent\textsuperscript{*} \\
    \scriptsize AIcrowd, Switzerland\\
    \scriptsize \texttt{florian@aicrowd.com}
\and
    Manuel Schneider\thanks{These authors have equal contribution} \\
    \scriptsize ETH Zurich, Switzerland\\
    \scriptsize \texttt{manuel.schneider@hest.ethz.ch}
\and
    Christian Scheller \\
    \scriptsize FHNW, Switzerland, \\
    \scriptsize AIcrowd, Switzerland
\and
    Jeremy Watson \\
    \scriptsize AIcrowd, Switzerland
\and
    Jiaoyang Li \\
    \scriptsize University of Southern California, \\ 
    \scriptsize United States of America
\and
    Zhe Chen \\
    \scriptsize Monash University, Australia
\and
    Yi Zheng \\
    \scriptsize University of Southern California, \\ 
    \scriptsize United States of America
\and
    Shao-Hung Chan \\
    \scriptsize University of Southern California, \\ 
    \scriptsize United States of America
\and
    Konstantin Makhnev \\
    \scriptsize HSE University, Russian Federation
\and
    Oleg Svidchenko \\
    \scriptsize HSE University, Russian Federation, \\ 
    \scriptsize JetBrains Research, Russian Federation
\and
    Vladimir Egorov \\
    \scriptsize HSE University, Russian Federation, \\ 
    \scriptsize JetBrains Research, Russian Federation
\and
    Dmitry Ivanov \\
    \scriptsize HSE University, Russian Federation, \\ 
    \scriptsize JetBrains Research, Russian Federation
\and
    Aleksei Shpilman \\
    \scriptsize HSE University, Russian Federation, \\ 
    \scriptsize JetBrains Research, Russian Federation
\and
    Evgenija Spirovska \\
    \scriptsize Netcetera, Switzerland
\and
    Oliver Tanevski \\
    \scriptsize Netcetera, Switzerland
\and
    Aleksandar Nikov \\
    \scriptsize Netcetera, Switzerland
\and
    Ramon Grunder \\
    \scriptsize Netcetera, Switzerland
\and
    David Galevski \\
    \scriptsize Netcetera, Switzerland
\and
    Jakov Mitrovski \\
    \scriptsize Netcetera, Switzerland
\and
    Guillaume Sartoretti \\
    \scriptsize National University of Singapore, \\
    \scriptsize Singapore
\and
    Zhiyao Luo \\
    \scriptsize National University of Singapore, \\
    \scriptsize Singapore
\and
    Mehul Damani \\
    \scriptsize National University of Singapore, \\
    \scriptsize Singapore
\and
    Nilabha Bhattacharya \\
    \scriptsize AIcrowd, Switzerland
\and
    Shivam Agarwal \\
    \scriptsize University of Duisburg-Essen, \\
    \scriptsize Germany
\and
    Adrian Egli \\
    \scriptsize SBB CFF FFS, Switzerland
\and
    Erik Nygren \\
    \scriptsize SBB CFF FFS, Switzerland \\
    \scriptsize \texttt{erik.nygren@sbb.ch}
    \and
    Sharada Mohanty \\
    \scriptsize AIcrowd, Switzerland \\
    \scriptsize \texttt{mohanty@aicrowd.com}
}

\date{}

\begin{document}
\maketitle
\newpage
\begin{abstract}
The Flatland competition aimed at finding novel approaches to solve the vehicle re-scheduling problem (VRSP).
The VRSP is concerned with scheduling trips in traffic networks and the re-scheduling of vehicles when disruptions occur, for example the breakdown of a vehicle.
While solving the VRSP in various settings has been an active area in operations research (OR) for decades, the ever-growing complexity of modern railway networks makes dynamic real-time scheduling of traffic virtually impossible.
Recently, multi-agent reinforcement learning (MARL) has successfully tackled challenging tasks where many agents need to be coordinated, such as multiplayer video games.
However, the coordination of hundreds of agents in a real-life setting like a railway network remains challenging and the Flatland environment used for the competition models these real-world properties in a simplified manner.
Submissions had to bring as many trains (agents) to their target stations in as little time as possible.
While the best submissions were in the OR category, participants found many promising MARL approaches.
Using both centralized and decentralized learning based approaches, top submissions used graph representations of the environment to construct tree-based observations.
Further, different coordination mechanisms were implemented, such as communication and prioritization between agents.
This paper presents the competition setup, four outstanding solutions to the competition, and a cross-comparison between them.
\end{abstract}


\section{Introduction}
\label{sec:intro}

Modern railway networks such as the one operated by the Swiss Federal Railways Company (SBB) are becoming increasingly complex, and the efficient coordination of the traffic on the network poses a significant challenge.
In a system that registers more than 10,000 train runs a day\footnote{\url{https://reporting.sbb.ch/verkehr?years=0,1,4,5,6,7&scroll=3135}}, even small disturbances such as a train malfunction can have a huge impact on the service quality and stability of the entire network.
Therefore, not only the initial scheduling but also the efficient, dynamic rescheduling of trains is essential.

The challenge of scheduling vehicles dates back several decades and was formally expressed as the ``vehicle scheduling problem'' (VSP)~\citep{bodin_classification_1981}.
Finding solutions to the VSP has been an active area of operations research (OR) ever since~\citep{foster1976integer,potvin1993parallel}.
Due to the increasing complexity and the need for efficient rescheduling in case of disruption, \citet{li2007vehicle} proposed an extension of the VSP to the broader ``vehicle re-scheduling problem'' (VRSP) which takes disruptions like vehicle breakdowns into account and represents a dynamic version of the VSP.

However, solving the VRSP incorporating all aspects given in a real world railway network is an NP-complete problem and does not allow for fast experimentation of automated traffic management, let alone the real-time rescheduling of trains.
Therefore, the research team at SBB started to explore new approaches, including multi-agent reinforcement learning (MARL)~\citep{gtc_USA_2018}.

In recent years, MARL algorithms have achieved remarkable successes on challenging multiplayer video game benchmarks such as StarCraft II~\citep{Vinyals2019,samvelyan2019starcraft}, Dota 2~\citep{openai2019dota}, hide-and-seek~\citep{baker2020emergent} and Capture the Flag~\citep{Jaderberg859}.
Besides idealized video game environments, cooperative multi-agent reinforcement learning also shows promise for many real-life applications such as network traffic signal control~\citep{arel2010reinforcement,tan2020coop} and real-time bidding~\citep{jin2018realtime}.
However, scalability to large number of agents, partial observability and communication constraints of individual agents remain major challenges.
Many recent approaches address these by learning decentralized policies in a centralized manner \citep{DIAL,Gupta2017,rashid2018qmix}.
With decentralized policies, communication becomes important.
\citet{DIAL}, \citet{CommNet}, \citet{ATOC} and \citet{TarMAC} addressed this by explicitly modelling communication between agents. 

To cope with large joint action spaces, value function decomposition methods learn centralized but factored global Q-functions, which handle coordination dependencies implicitly \citep{Sunehag2018VDN,rashid2018qmix,Son19QTRAN,Wang2020LearningNDV,wang2021qplex,rashid2020weighted,wang2021rode}.
\citet{COMA} and \citet{Lowe2017MADDPG} proposed the paradigm of centralized critic with decentralized actors for multi-agent policy gradient algorithms.
Extending these works, \citet{iqbal2019AAC} proposed a shared attention mechanism and \citet{wen2019probabilistic} introduced a recursive reasoning mechanism.

To address the computational complexity of a full system simulation and in light of the recent development in reinforcement learning (RL) research, SBB in collaboration with AIcrowd, developed a framework that provides a simplified yet representative environment to study dynamic train (re-)scheduling~\citep{mohanty2020flatlandrl}.
A first edition of the Flatland competition was held in 2019 with the aim of discovering novel approaches to the VRSP with a special emphasis on RL-based solutions.

In this paper, we present the Flatland competition run at NeurIPS 2020 and the main insights gained from it.
We outline the competition format in the next section.
We then present common approaches taken by top submissions in \sectionref{sec:methods} and proceed with detailed descriptions of outstanding solutions – provided by the teams themselves – in \sectionref{sec:solutions}.
In \sectionref{sec:results}, we present a comparative analysis of these solutions, and discuss our findings in the final section.

\section{Competition}
\label{sec:competition}

The competition intended to foster research into novel solutions to the real-world VRSP problem given by a modern railway network.
Two competition tracks were available for submitting either RL or more established OR formulations to encourage cross-pollination of ideas.
The competition was organized on the AIcrowd platform and sponsored by the Swiss Federal Railway Company (SBB), Deutsche Bahn (DB) and Société Nationale des Chemins de fer Français (SNCF).

Entry barriers were low to encourage participation and allow teams from diverse backgrounds and expertise to contribute.
Participants could use a Starter Kit\footnote{\url{https://gitlab.aicrowd.com/flatland/neurips2020-flatland-starter-kit}}, which could be directly submitted to the competition, to support experimentation with RL approaches.
Additionally, participants had access to advanced baselines\footnote{\url{https://gitlab.aicrowd.com/flatland/neurips2020-flatland-baselines}} based on the RLlib framework~\citep{liang2018rllib} with brief documentation outlining the ideas behind each baseline as well as experimental results (see \appendixref{app:baselines} for details).

\begin{figure}[t]
    \centering
    \includegraphics[width=\linewidth]{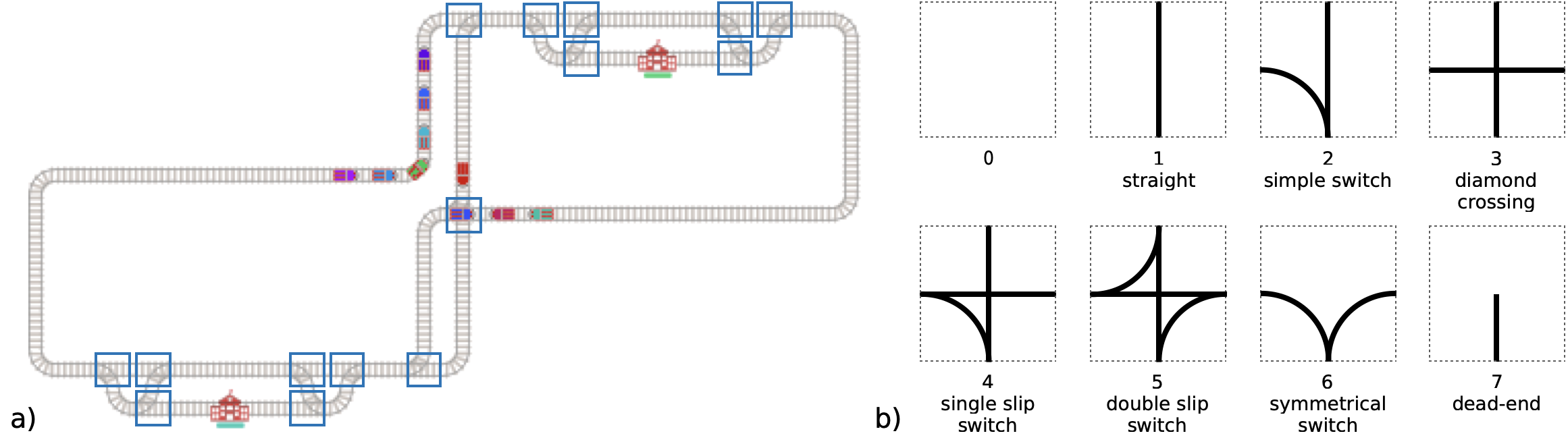}
    \caption{\textbf{Flatland environment.} a) Visualisation of a Flatland environment. The agents (colored trains) move on tracks (grey) and have to reach their targets (stations). Switches allow the trains to change tracks (blue rectangles). b) In Flatland, a cell has one of these eight rail configurations, a mirror image of or a rotation of 90\textdegree, 180\textdegree\ or 270\textdegree\ from it. Case 1 can also be curved.}
    \label{fig:environment}
\end{figure}

\subsection{Environment}
The Flatland competition was built on the Flatland framework~\citep{mohanty2020flatlandrl} that provided randomly generated environments for the agents to operate in (see \figureref{fig:environment}).
A Flatland environment contained a $w \times h$ rectangular grid of cells where some cells contained ``rails''.
Rails were either straight or curved at 90\textdegree \space and linked cells that were adjacent horizontally, vertically or both.
The rails in some cells formed ``switches'', which joined 3 or all 4 adjacent cells in different ways.
Available choices for an agent representing a train arriving at a switch were dependent on the direction of entry into a cell, known as ``transitions'' from an entry direction to an exit direction.
The possible transitions in Flatland are shown in \figureref{fig:environment}.
For a given entry direction, the track layout only ever allowed for 1 or 2 choices of exit direction.
In the environments of the competition, all cells were part of at least one track cycle (or at least two cycles when counting the directions separately), and no ``dead-ends'' existed.

\subsection{Agents}
A given environment contained $n$ agents $a_i, i \in \{1, \ldots, n\}$.
Agents were assigned a starting cell (origin), a direction $d_{i} \in \{N,E,S,W\}$ available at the origin, and a target cell.
Thereby, multiple agents could share the same origin (agents started an episode ``off the grid''), the same target or both, but origins were distinct from target cells.
The origin and target cells were simple rails, not switches, and could be reached from both directions.
For each timestep of an episode, every agent issued one of five actions: go forward (\texttt{MOVE\_FORWARD}); select a left turn (\texttt{MOVE\_LEFT}); select a right turn (\texttt{MOVE\_RIGHT}); halt on current cell, always valid (\texttt{STOP}); and no-op, always valid (\texttt{DO\_NOTHING}).
If an agent chose the no-op action (\texttt{DO\_NOTHING}), it carried on as before, either off the grid, stationary in a cell, or moving forward.
An agent that did not explicitly select an action would perform a no-op.
If an agent selected an action that was not currently valid, the action failed which was equivalent to performing a no-op.
Agents only entered the grid when they issued an action in \{\texttt{MOVE\_FORWARD}, \texttt{MOVE\_LEFT}, \texttt{MOVE\_RIGHT}\} which was carried out during the timestep of entry.
An agent could choose to stay off the grid for the whole episode.
If an agent's origin was not empty when the agent attempted to enter the grid, the action failed.
Agents could only move forward (including left and right turns) or stop, but they couldn't reverse in situ.
Thus, if two agents ``collided'' while heading in opposite directions, they became deadlocked for the rest of the episode, and could not reach their targets.
When moving, agents proceeded one cell per timestep, and all actions were enacted at once.
Once an agent reached its target, it permanently disappeared from the grid and no longer occupied a cell. 

An agent could also malfunction, in which case the agent's actions had no effect and the agent could not move until the malfunction was resolved.
Each agent randomly entered a malfunction state with a fixed probability given by a malfunction rate which was set for an environment and shared between all agents.
The malfunction rate varied between 0, i.e. no malfunctions, and 0.004 meaning one malfunction was expected every 250 time steps on average.
When an agent malfunctioned, the malfunction duration, i.e., the number of timesteps to be spent in the failed state, was drawn from a discrete uniform distribution in $[20,50] \cap \mathbb{N} $.
The remaining duration of the malfunction was available to the agent.
However, the malfunction rates and duration ranges were not available to the agents at any time.

\subsection{Task}
The task of the competition was to bring as many agents to their targets in as little timesteps as possible.
In the RL formulation, the agents received a penalty of -1 for each timestep they did not attain their targets, whether they were on or off the grid, moving or stationary.
After an agent reached its target, the reward was 0 for each remaining timestep.
The episode was terminated when all agents reached their targets or when the timestep limit
\begin{equation*}
    t_{max} = \left \lfloor{ 8 \left(w + h + \frac{n}{c} \right)} \right \rfloor
\end{equation*}
was reached, with $n$ being the number of agents and $c$ the number of cities in the environment (see \appendixref{app:env_configs}).
The sum $w + h$ was a rough estimate for an agent's shortest path, and the fraction $n / c$ an estimate for the average number of agents that started after each other.
Every agent received an additional reward of 1 if all the agents had reached their targets.
Thus the total episode reward for each agent $a_i$ was a negative integer $t_{a_i} \in \{ t \in \mathbb{N}^{-} | t \geq -t_{max} \}$.
The intention was that agents act cooperatively, so the overall normalized score for an episode combined the $n$ agents' individual rewards:
\begin{equation*}
    s := 1 + \frac{\sum_{i=1}^n t_{a_i}}{n \cdot t_{max}}  \in [0,1) \quad .
\end{equation*}

The participants had to solve as many environments as possible within an 8 hour time limit.
The overall score $S$ of the submission was given by the sum of the normalised environment scores $S := \sum_j^m s_j$ with $m$ being the number of environments solved.

In contrast with common RL environments, Flatland allowed access to its full internal state.
An integral part of the challenge was to design observation spaces that RL algorithms could learn from.
This was challenging due to the varying number of cells and agents in the grid, and due to the complex dynamics of the environment.

\subsection{Evaluation}
Once implemented, participants submitted their solutions, i.e. code and any trained model data, to the AIcrowd evaluator.
A competition submission was labelled by the submitter as `RL', `OR' or `other', dependent on the chosen approach.
The evaluator tested the performance of the submitted solutions on a fixed set of pre-generated test environments, which were not disclosed to the participants.
For every test $T_k, k \geq 0$, ten environments $E_{k,l}, l \in \{0, \ldots, 9\}$ were generated.
The environments of the same test shared the same parameters, except for the malfunction rate depending on $l$, while the environments' complexity progressively increased from test to test (see \appendixref{app:env_configs}).
As the participants had to solve as many environments as possible, enough environments were provided to exceed the anticipated capabilities of all participants.

Timeouts were enforced during the evaluation: 10 minutes per environment for initial planning, then 10 seconds were allowed per timestep. The submissions could use up to 4 CPU cores (see \appendixref{app:eval_setup} for details).
If a timeout was triggered, that environment scored 0.
The evaluation ended when one of the following conditions was met: (a) the evaluator encountered 10 consecutive timeouts; (b) fewer than 25\% of the agents reached their target during a test; (c) the maximum evaluation duration of 8 hours had passed.

A set of 2 example environments per test was made available to the participants,\footnote{\url{https://www.aicrowd.com/challenges/neurips-2020-flatland-challenge/dataset_files}} along with the parameters of all environments.\footnote{\url{https://flatland.aicrowd.com/getting-started/environment-configurations.html}}
Participants were also free to use the environment generator to create their own training sets.
A set of ``expert'' solutions generated using the winning solution from the 2019 Flatland competition was also made available to participants.

\section{Methods}
\label{sec:methods}

Participants explored a variety of ways to solve the problem posed by the NeurIPS 2020 Flatland competition.
Overall, the solutions were categorised into RL, OR and mixed approaches. We describe the methods used by four outstanding solutions of the Flatland competition, three RL solutions and one OR.

Multiple teams approached the competition task from the perspective of \emph{Multi-Agent Path Finding} (MAPF)~\citep{SternSOCS19}. 
The input of a classic MAPF problem is an unweighted graph and a set of agents, each with a start and target vertex (see \appendixref{app:graph} for more information on the graph representation of Flatland).
At each timestep $t$, an agent either moves to an adjacent vertex or waits at its current vertex.
Agents are at their start vertex at $t=0$ and remain at their target vertex after they completed their paths.
The task is to move all agents to their target vertices without collisions within finite timesteps while minimizing the sum of their travel times.
However, unlike classic MAPF problems, Flatland poses additional challenges in the form of a highly constrained and irreversible environment and a sparsity of important decisions.
Despite the differences to MAPF, there are similarities to some MAPF variants: (i) Trains enter the environment over time and leave it after reaching their target stations, which is related to online MAPF~\citep{SvancaraAAAI19,PRIMAL2}; (ii) As many trains as possible (instead of all trains) should reach their target stations before a given timestep, which is related to MAPF with deadlines~\citep{MaIJCAI18}; (iii) Trains breakdown randomly while moving and are then stationary at the breakdown location for a number of timesteps, which is related to MAPF with delay probabilities~\citep{MaAAAI17,AtzmonICAPS20}.

The winning solution fell into the OR category and successfully applied a MAPF approach that combines various MAPF algorithms and optimization techniques.

The best-ranked RL teams used diverse reinforcement learning algorithms, however they all settled on a common set of approaches which we will summarise briefly (see \appendixref{app:solution_configs} for specifications of the RL solutions).

\paragraph{Tree Observations}
Most RL solutions used tree-based observations. Similarly to the OR approach, these observations leverage the underlying graph structure of the railway network. They are generated by spanning a binary tree from the current position of each agent, with branches following the allowed transitions until some maximum depth is reached. The resulting observation contains the information gathered at the switches along each branch. Various features can be recorded such as the presence of the agent's target on the current branch, the presence of other agents, the length of the shortest path from the current cell to the agent's target, etc.
The winning RL teams used trees with depth between 1 and 3. They all experimented with various features and settled on sets containing between 4 and 11 features. This involved a trade-off between the additional information brought by each feature and the time taken to calculate them at each timestep.

\paragraph{Reward shaping}
Since the rewards provided by the Flatland environment were sparse, participants experimented with various alternative formulations to help the learning process.
While reward shaping can greatly help, it needs to be done carefully to prevent the onset of undesirable agent behavior. If the penalization for unwanted behavior was too strict, the agents could learn that the best course of action was to never enter the grid. Conversely, a large positive reward for a sub-goal could incite agents to exploit it, resulting in agents that would never reach their targets. 
The winning RL solutions all used bonuses for reaching the target, penalties for deadlocks, and additional smaller rewards for custom circumstances.

\paragraph{Maximum Occupancy}
The number of agents in the grid increases the probability of deadlocks and of agents malfunctioning. One way to avoid this is to limit the maximum occupancy of the environment.
The winning teams all used some metrics or heuristics to limit the maximum number of concurrent agents.

\paragraph{State Masking}
\label{sec:masking}
The rail grid cells can be classified into three categories (see also \figureref{fig:cell_types} in \appendixref{app:cell_types}): (a) non-decision cell, i.e., there is no interesting decision in this cell or its neighbouring cells; (b) stopping cell, i.e., there is a neighbouring cell from which multiple ($\geq$3) transitions are possible; (c) decision cell, i.e., there are multiple transitions available at the current cell.
The majority of cells in Flatland are non-decision cells, in which no significant decision can be made.
All winning solutions masked non-decision cells in order to improve performance either at training time, at evaluation time, or both.

\section{Outstanding Solutions}
\label{sec:solutions}

In this section, we detail the approaches and results of four outstanding solutions, one OR and three RL, to the NeurIPS 2020 Flatland competition.
We placed the focus on RL solutions to support further research into the still experimental MARL approaches to the VRSP.


\subsection{Multi-Agent Path Finding for Large-Scale Rail Planning}

Team \textit{An\_Old\_Driver}: Jiaoyang Li, Zhe Chen, Yi Zheng, Shao-Hung Chan; 1st place OR and overall.


\subsubsection{Method}
The Flatland Challenge at its core is a MAPF problem. The first two differences discussed in \sectionref{sec:methods} can be addressed by small modifications of existing MAPF algorithms. The last difference, namely the malfunctions, can be handled during execution~\citep{MaAAAI17}. We therefore first plan collision-free paths under the assumption that no malfunctions occur in the beginning and then handle malfunctions once they occur.

\paragraph{Planning Collision-Free Paths}
We use Prioritized Planning (PP)~\citep{WHCA}, a simple but widely-used MAPF algorithm to generate the initial collision-free paths for all trains.
PP first sorts all trains in a priority ordering and then, from the highest priority to the lowest priority, plans a shortest path for each train while avoiding collisions with the already planned paths of higher-priority trains.
For efficiency, we use \emph{Safe Interval Path Planing} (SIPP)~\citep{SIPP} instead of A* to plan each such path in PP, since it avoids temporal symmetries. 

Although PP can find collision-free paths rapidly, its solution quality is far from optimal.
We therefore use Large Neighborhood Search (LNS)~\citep{LNS} to improve the solution quality. 
We follow \citet{LiAAMAS21} by using PP to generate an initial solution and repeating a neighborhood search process to improve the solution quality until the iteration limit is reached. In each iteration, we select a subset of trains and replan their paths using PP. The new paths need to avoid collisions with each other and with the paths of other trains. We adopt the new paths if they result in a smaller sum of the travel times.
As the Flatland Challenge provides 4 CPUs for evaluation, we run 4 LNS in parallel in practice. As we are given an overall runtime limit of 8 hours, we use simulated annealing to decide the iteration limit for each environment.

Although SIPP runs significantly faster than A*, it is still slow when there are thousands of trains because, as the paths of more trains are planned, SIPP has to plan paths that avoid collisions with more existing paths, resulting in its runtime growing rapidly.
We therefore propose a \emph{lazy planning} scheme where we plan paths only for some of the trains in the beginning, then let the trains move, and plan paths for the rest of trains during execution. 
Although delaying the planning time of the trains may delay their departure times, which in turn may delay their arrival times, lazy planning has two implicit benefits: (1) it avoids pushing too many trains to the environment at once, which could potentially prevent severe traffic congestion; and (2) when planning paths during execution, we take into account the influence of the malfunctions that have already happened or are happening.

\paragraph{Recovering from Malfunctions}

When a train encounters a malfunction during execution, deadlocks could happen if the trains stick to their original paths.
\emph{Minimum Communication Policies} (MCP)~\citep{MaAAAI17} avoids the deadlocks by stopping some trains to maintain the ordering that each train visits each location. It guarantees that all trains can reach their target stations within a finite number of timesteps. 
However, MCP sometimes may stop trains unnecessarily. We therefore develop a \emph{partial replanning} mechanism to avoid such unnecessary waits. When train A encounters a malfunction at some timestep, we collect all switch and crossing rail segments that train A is going to visit in the future and then collect the trains who are going to visit at least one of these rail segments after train A. We replan the paths of these trains in the prioritized planning manner and terminate when either new paths are planned for all of these trains or the runtime limit is reached.

\subsubsection{Results}

Our solution is implemented in C++ and is available from our public repository.\footnote{\url{https://github.com/Jiaoyang-Li/Flatland}} Our basic solution, namely PP with A* (instead of SIPP) plus MCP, solved 349 instances within 8 hours with a score of 282.56. When we replace A* with SIPP, our solution solved 3 more instances with a score of 285.4. LNS then improved the score to 289.1, and partial replanning further improved it to 291.873. Eventually, with the help of lazy planning, we solved 362 instances and reached our highest score of 297.5.
More details can be found in \citep{LiICAPS21}.

\subsection{PPO with Communication and Departure Schedule}
\label{sec:rl1}

Team \textit{JBR\_HSE}: Konstantin Makhnev, Oleg Svidchenko, Vladimir Egorov, Dmitry Ivanov, Aleksei Shpilman; 1st place RL.


\subsubsection{Method}

Our solution is based on Proximal Policy Optimization (PPO) \citep{PPO} enhanced with between-agent communication.
Communication has been shown to help agents to solve simple coordination problems \citep{CommNet,TarMAC}, and our solution extends this result to the complex Flatland environment.

\paragraph{Rewards} 

For each agent, its reward is defined as $0.01 \times \Delta d - 5 \times \texttt{is\_deadlocked} + 10 \times \texttt{has\_finished}$, where $\Delta d$ is the difference between the minimal distances to the destination point from its previous and current positions, and \texttt{is\_deadlocked} and \texttt{has\_finished} are the respective indicators of whether the agent is deadlocked or has successfully completed the episode.

\paragraph{Architecture}

In our solution, the agents are based on the actor-critic framework.
The actor makes decisions whenever encountering a switch, based on features from three sources (see \figureref{fig:architecture} in \appendixref{app:architecture}):

First, it receives information about the agent itself, such as its position on the map and indicators of whether it is deadlocked or malfunctioning. At the beginning of the episode, each agent is also assigned with a handle, a random number uniformly sampled from $[0, 1]$ which is supposed to represent priority. This information is processed with a fully-connected network (denoted as ``Common Features Net'').

Second, the actor observes the most relevant part of the map, encoded as the edges of a tree with a depth limited to 3. The features of each edge include its length, distance to the destination from its nodes, as well as information about other agents, including their positions and handles. The features of different edges are combined with a recursive network (denoted as ``Tree Features Net''), albeit replacing it with a fully-connected network produces similar performance.

Third, prior to action selection, the actor receives a set of floating-point vector messages generated by the neighbouring agents, i.e. the agents observed on the tree. Similarly to ATOC \citep{ATOC}, the messages are processed with self-attention layers \citep{attention}.

All components are trained end-to-end using PPO. The agents share experiences and network parameters. The architecture of the critic is similar to that of the actor, except the critic does not condition on communication signals. For a more complete description of the architecture and the features, please refer to our code.\footnote{\url{https://github.com/jbr-ai-labs/NeurIPS2020-Flatland-Competition-Solution}}

\paragraph{Departure Schedule}

To limit the maximum occupancy, we introduce a departure schedule. After pre-training agents without the schedule, we train a supervised classifier that determines the chances that each agent reaches its destination. Then, we only allow each train to start moving if the predicted probability exceeds a threshold, set to 0.92 at the start of the episode, and gradually lowered over time.

\subsubsection{Results}

We found that a carefully tuned PPO with simple heuristics for launching agents can achieve results comparable to simple algorithms from the operations research field while being quite fast in execution. Among the modifications that we applied to the classic PPO algorithm, communication was by far the most impactful. It significantly increased the arrival rate without compromising scalability. One particular limitation of our solution is that we train the agents in relatively small environments. Consequently, their performance may degrade as the environment scales up.

\subsection{RL with Centralized Priority Assignment}
\label{sec:rl2}

Team \textit{Netcetera}: Evgenija Spirovska, Oliver Tanevski, Aleksandar Nikov, Ramon Grunder, David Galevski, Jakov Mitrovski; 2nd place RL.


\subsubsection{Method}
We designed an RL solution, where at each timestep each train generates an experience. We use centralized training of an Ape-X model as described in \cite{horgan2018distributed}. In the following subsections a detailed explanation for all parts of the solution is presented.

\paragraph{Representation of the Environment}
In order to optimize the calculations, we represent the environment as a graph, where each vertex represents a switch and possible arrival directions to the switch. More formally, a vertex is represented as a state-direction pair $(s,d)$. A directed edge between two vertices in the graph $(s_{1},d_{1})$ and $(s_{2},d_{2})$ exists if there is a directed path in the environment from $s1$ to $s2$, such that the directions of the train on switches match $d_{1}$ and $d_{2}$ correspondingly. A representation of a simple environment is given in \figureref{fig:graph} in \appendixref{app:graph}.

\paragraph{Representation of the State}
The state of the agent should describe its immediate surroundings well, so it can make an intelligent and informed choice for the next action. The best choice of features was obtained experimentally. Our best state representation uses a tree observation of depth 1 which includes features about the shortest path, the status and the priority of the agent and conflict information (for example: deadlock, deadlock in the next whole segment until the next usable switch). During the course of the competition we experimented with many other features, like statistical information about the path and the environment, time tick info (the number of elapsed timestamps since the beginning of the episode), info about following switches and other agents. These features proved to only create noise in the model and reduced its performance.

\paragraph{Priority}
Our model's most important feature, which led to significant improvement in its performance, was priority. The concept of priority is used for solving conflicts between agents, which happen on a regular basis during simulation. We build an undirected conflict graph, where each vertex represents an agent and the edges in the graph represent the conflicts between agents. Assigning priority to the agents is a graph coloring problem, where two neighboring vertices cannot have the same level of priority at the same time. 

We have tried several heuristics for choosing the agents that have priority. The best one was to give the highest priority to the vertex with highest degree. The logic behind this heuristic is that the agent that causes the largest number of conflicts will go first, freeing the path for most agents in the next steps.

\paragraph{Reward}
In our final reward setting, the only positive reward was reaching the target. We also included smaller and larger penalization for correcting unwanted behaviors. The biggest negative penalization was for getting into deadlock. With smaller penalization we correct not following priority, missing the shortest path, unwanted stopping, etc. 

\paragraph{Models}
Although we experimented with on-policy and off-policy models, the best results were achieved with Ape-X model. We used the RLlib library \cite{liang2018rllib} for model implementations.
We introduced the concept of curriculum learning~\citep{Bengio2009}, meaning the difficulty of the provided training environments changed over time. The idea was for the agents to learn simple behaviors first. As the training progressed, so did the complexity of the training environments, allowing the agents to learn more complex behaviors.

\subsubsection{Results}
Our solution\footnote{\url{https://github.com/netceteragroup/Flatland-Challenge}} was able to solve 228 environments, where the most complex one consisted of 181 agents and 20 cities. 88.1 \% of the agents were able to reach their target destination. Our final score was 181.5 and the reason for termination was that we reached overall the time limit of 8 hours per solution.
Our solution performs well in avoiding deadlocks. For the smaller environments, all the trains are able to reach their targets. The main disadvantage of our solution is that it is too `cautious', meaning that for really large environments more trains should be on the map simultaneously.

\subsection{TrainfficLight: Decentralized RL for MAPF}
\label{sec:rl4}

Team \textit{MARMot-Lab-NUS}: Guillaume Sartoretti, Zhiyao Luo, Mehul Damani; 4th place RL.

\subsubsection{Method}

Our solution builds upon the multi-agent decentralized RL framework for MAPF in grid environments proposed in our previous work, PRIMAL~\cite{sartoretti2019primal}. Taking inspiration from PRIMAL, we construct a rich feature based observation for agents and use a shallow fully connected neural network architecture. We use the popular A3C learning algorithm for training~\cite{mnih2016asynchronous}. We also draw  inspiration from the use of traffic lights, which are effective in managing traffic at bottlenecks such as junctions and crossings. Despite the traffic-light-like mechanism we propose, our solution can still be implemented in a fully decentralized manner, by relying on local communication between nearby agents only.

\paragraph{State Masking}
\label{section:state_masking}

We classified the rail grid cells as described in \sectionref{sec:masking}.
Since the majority of cells in Flatland are non-decision cells and in these cells, we hard code a \texttt{MOVE\_FORWARD} action or a \texttt{STOP} action, which is determined solely by the occupancy of the single reachable neighbouring cell to ensure safety.
Correspondingly, we remove the state-action pairs of non-decision cells from training.
An obvious advantage of the state classification is that we drastically reduce the number of samples agents needed to learn in an episode.

\paragraph{Observation}

According to our state masking, decisions are only made at stopping points, i.e., a state before agents finalize their headings on a junction. The observation of an agent is tree-shaped but with a relatively shallow depth (only immediate information about the most immediate junction). Specifically, it contains a series of handcrafted and hand-selected features on the future junction (i.e., turn left, go forward and turn right). Each choices consists of 9 scalar values, including 1) whether current heading has solution to the goal 2) optimal path length to its goal, 3) number of agents blocking with in the next junction, 4) total number of agents blocking along the optimal path, 5) number of agents crashed along the optimal path, 6) total number of agents blocking on junctions along the optimal path, 7) number of agents queuing within the next junction, 8) distance to the next junction and 9) whether the closest stopping point is occupied. In total, each agent's observation is a $3 \times 9$ vector of these features, ultimately represented as a $1 \times 27$ vector.

\paragraph{Clusters}

In our solution, the real bottlenecks are not the low-level individual junction cells or switches where agents have multiple possible decisions, but are the closely connected clusters of these cells which need to be navigated effectively.
We formally define clusters as a collection of connected grid cells where each cell has at least three possible transitions (see \figureref{fig:cell_types} in \appendixref{app:cell_types}).
Since they usually consist of multiple decision cells, clusters also negatively affect tree observations by occupying a significant depth of the tree.
This makes the observation highly localized which affects the ability of agents to do long-term planning.

\paragraph{Traffic Lights}

In order to effectively resolve the bottleneck created by clusters, we propose a smart traffic light inspired mechanism that controls entry into these clusters by only allowing a single agent to occupy the cluster at a time. 
When an agent is currently occupying the cluster, the traffic light signal is red for all entry cells into the cluster. 
When there is no agent occupying the cluster, then the traffic light signal switches to green for the entry cells into the cluster which currently have an agent waiting. 
Although the implementation of this mechanism is handcrafted in our work, we believe that it would be easy to train a RL-based controller for the traffic light policy, which could make the system even more robust. This will be investigated in future works.

\paragraph{Departure Initialization}

We use two simple rules for agent initialization. 
First, we set a soft upper bound for the number of agents in the system which is a function of the grid size. 
Second, we initialize agents having the same start and end locations one after the other.
These agents generally tend to follow the same trajectory in a line and as a result, the chances of conflict between them are minimal. 

\subsubsection{Results}

Our solution\footnote{\url{https://github.com/marmotlab/flatland-challenge-neurips-2020}} achieved a score of 127.9 and was able to plan successfully for 67\% of the agents. 
The most effective component of the solution was the introduction of traffic lights, which helped prevent deadlocks and conflicts in clusters. 
Our solution, being reactive, was also relatively unaffected by malfunctions. 
We also found that departure initialization using the heuristics described above has a considerable impact on performance.

\section{Competition Results}
\label{sec:results}

\begin{table}[ht]
    \centering
    \begin{tabular}{|l|l|l|l|l|}
        \hline
        \textbf{Team} & \textbf{Rank} & \textbf{Score} & \textbf{\% Trains Done} & \textbf{\# Environments} \\ \hline
        An\_Old\_Driver & OR 1 & 297.507 & 98.6\% & 363 \\ \hline
        JBR\_HSE & RL 1 & 214.150 & 78.5\% & 336 \\ \hline
        Netcetera & RL 2 & 181.497 & 88.1\% & 229 \\ \hline
        MARMot-Lab-NUS & RL 4 & 127.912 & 66.4\% & 230 \\ \hline
    \end{tabular}
    \caption{\textbf{Competition results.} Overall scores, completion rates and number of tackled environments of the four solutions presented in this paper.}
    \label{tab:evaluation}
\end{table}

We compared the competition performances of the four solutions presented in the previous section.
\tableref{tab:evaluation} shows their overall scores, completion percentages and number of encountered environments.
The OR solution (An\_Old\_Driver) tackled the most environments with a total of 363, followed by the winning RL solution (JBR\_HSE) with 336.
Interestingly, the second RL solution (Netcetera) tackled 1 environment less (229) than the fourth RL solution (MARMot-Lab-NUS, 230), but scored better on average resulting in a better ranking.
This perfectly illustrates the two competing goals of the challenge: maximizing the number of agents that reach their targets vs minimising the time needed to bring the agents there.

\begin{figure}[ht]
    \centering
    \includegraphics[width=0.9\linewidth]{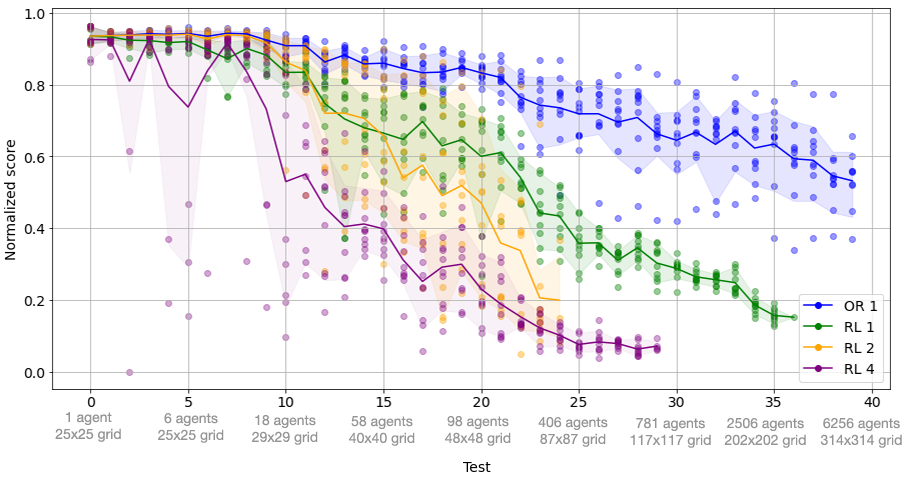}
    \caption{\textbf{Normalized environment scores.} The dots indicate the normalized scores the solutions achieved for a given environment. The lines represent the mean of the 10 environment scores per test (x-axis) and the filled area the 0.1-quantile.}
    \label{fig:evaluation}
\end{figure}

\figureref{fig:evaluation} shows the normalized scores of the four solutions after 90 hour of evaluation, without enforcing the termination conditions of the competition (see \appendixref{app:completation} for the completion rates).
For simple environments and a small number of agents, all four solutions received a very high score with some exceptions for the fourth RL solution.
Scores start to decline around test 10 when the number of additional agents per environment increased faster (+8 agents per test) than during the 10 tests before (+1 agent).
While the fourth RL solution had some outliers in the earlier tests, the second RL solution had a higher variance later on.
The best RL solution and the OR solution show comparably less spread of the scores which indicates that they could better adapt to all sorts of environments.
A reason for the best RL solution to cope better with the larger number of agents compared to the other RL solutions could be given by the learned communication between agents, which was responsible for better performances during the experimentation phase by the team (see \sectionref{sec:rl1}).

All three RL solutions used different learning algorithms (PPO, Ape-X, A3C) but, at first glance, there seems to be no performance difference based on the algorithm choice which is consistent with the results from the Flatland baselines~\citep{mohanty2020flatlandrl}.

It is also worth highlighting that both centralized and decentralized approaches were effective and had comparable performance.
An\_Old\_Driver used an OR-based priority-planning approach to centrally plan collision-free paths.
In contrast, all RL solutions were based on the centralized training with decentralized execution paradigm, whereas the arising coordination problem was solved in either a centralized or decentralized manner.
JBR\_HSE supplemented observations based on a local tree observation and hand-crafted features with local learned communication.
Netcetera constructed a global conflict graph to assign priorities in a centralized fashion during execution to solve conflicts between agents.
MARMot-Lab-NUS used a local tree observation in combination with decentralized traffic signal clusters that can be implemented by local communication around critical switches.

In parallel to the competition, special ``Community Prizes'' were available for contributions that supported the participants' experimentation.\footnote{\url{https://discourse.aicrowd.com/t/flatland-community-prize}}
The top community contributions\footnote{\url{https://discourse.aicrowd.com/t/neurips-2020-flatland-winners}} included innovative observation builders for RL, notebooks to leverage the RLlib baselines, and a tool to visually analyze the agents' behaviour (see \appendixref{app:visualization}), which won first place.

\section{Conclusion}
\label{sec:conclusions}

The results show that RL solutions are still a considerable distance away from OR based solutions.
OR solutions have the advantage that they can search the joint configuration space in a centralized manner, and, therefore, allow agents to exhibit coordinated maneuvers that can handle complex situations.
A single deadlock in Flatland can have a domino effect and further work needs to be done to address such events effectively.
Defining the right observations has proven to be difficult and more research is needed to investigate the influence of specific features.
For example, the local observation of an agent could be significantly enriched by combining a tree observation with a local grid observation to enable long-term planning and spatial reasoning in the region surrounding each agent.
Also network structures with a temporal component such as an LSTM could improve the long-term planning, but accommodating such an approach with state masking is challenging.
In the competition, heuristics customised to the specifics of Flatland and the task at hand, such as departure schedules and prioritisation, greatly improve the performance of solutions.
On the other hand, coordination mechanisms such as traffic light policies (see \sectionref{sec:rl4}) and the graph based prioritisation (see \sectionref{sec:rl2}) could potentially be improved by introducing RL-based controllers instead of manually implemented ones.
Overall, the Flatland competition 2020 demonstrated the potential of RL approaches but also showed that more systematic research is needed to find effective methods to solve the VRSP for problems such as Flatland.

\section*{Acknowledgements}
\label{sec:acknowledgements}
We would like to thank the competition sponsors Swiss Federal Railways Company (SBB), Deutsche Bahn (DB) and Société Nationale des Chemins de fer Français (SNCF) as well as all the people that helped with the planning, preparation and facilitation of the competition. Special thanks to Irene Sturm and Gereon Vienken from DB and François Ramond from SNCF. We are also grateful to Shivam Khandelwal, Jyotish Poonganam and Yoogottam Khandelwal from AIcrowd for their parts in the setup and maintenance of the competition.\newline
\textbf{An\_Old\_Driver:} We thank Han Zhang for initially trying some ideas. We thank our team advisors Danial Harabor, Peter J. Stuckey, Hang Ma, and Sven Koenig for helpful discussions and comments. The research at the University of Southern California was supported by the National Science Foundation (NSF) under grant numbers 1409987, 1724392, 1817189, 1837779, and 1935712 as well as a gift from Amazon. The views and conclusions contained in this document are those of the authors and should not be interpreted as representing the official policies, either expressed or implied, of the sponsoring organizations, agencies, or the U.S. government.\newline
\textbf{JBR\_HSE:} This research was supported in part through computational resources of HPC facilities at HSE University, Russian Federation. Support from the Basic Research Program of the National Research University Higher School of Economics is gratefully acknowledged.\newline
\textbf{Netcetera:} We thank Netcetera for the support for the participation in the Flatland challenge. We also thank Darko Filipovski, Nikola Velichkovski and Zafir Stojanovski that were part of the previous rounds of the challenge.


\appendix
\captionsetup{format=plain}
\counterwithin{figure}{section}
\counterwithin{table}{section}

\section{Starter Kit and Baselines}
\label{app:baselines}

Multiple ready-made solutions and advanced baselines were provided to participants to support quick experimentation.

\subsection{Starter Kit} The Starter Kit offered a short and readable implementation of the DDDQN algorithm~\citep{wang2016dueling} using PyTorch~\citep{paszke2019pytorch}. By default, it used a tree observation with 25 features and depth 2. It could be cloned from the AIcrowd repository and submitted directly to the challenge, granting a score of 54.646.
The Starter Kit was designed as a simple entry point for non-expert participants. A detailed documentation offered a walk-through of the code.\footnote{\url{https://flatland.aicrowd.com/getting-started/rl.html}}
It also included support for experiment logging and hyperparameter tuning using Weights \& Biases~\citep{wandb}.

\subsection{RLlib Baselines}

Baselines built on the RLlib framework~\citep{liang2018rllib} were also provided. RLlib is an open-source library for reinforcement learning that offers a unified API for various RL methods and is designed for scalability. The RLlib baselines included four reinforcement learning methods: Ape-X~\citep{horgan2018distributed}, PPO~\citep{PPO}, PPO with a centralized critic~\citep{iqbal2019AAC} and MARWIL~\citep{MARWIL}. They also included two improvements specifically designed for the Flatland environment: the ability to skip cells where no meaningful decision had to be taken, and the ability to mask invalid actions. By default, the baselines used a tree observation with 25 features and depth 2. An alternative observation that used the expected density in each cell was also available.

Some of the baselines used imitation learning (IL) approaches in order to leverage expert demonstrations. MARWIL uses demonstrations to learn a policy that can in theory exceed the performance of the expert policy. In addition, the Ape-X and PPO baselines could also be trained using a mixture of IL and RL experiences. Full details can be found in \cite{mohanty2020flatlandrl}.

\section{Environment Configurations for the Evaluation}
\label{app:env_configs}

The environments of the evaluation were generated with increasing difficulty. All environments were based on the same parameters:
\begin{align*}
    n\_envs\_run & := 10 \\
    min\_malfunction\_interval & := 250 \\
    max\_rails\_in\_city & := 4 \\
    malfunction\_duration & := [20,50] \\
    max\_rails\_between\_cities & := 2 \\
    speed\_ratios & := \{ 1.0: 1.0 \} \\
    grid\_mode & := False
\end{align*}
For every environment $l\in\{0, \ldots, 9\}$ of a test $k \geq 0$, the following parameters were calculated:
\begin{align*}
    malfunction\_interval_l & := l \cdot min\_malfunction\_interval \\
    n\_agents_{k+1} & := n\_agents_{k} +  \lceil 0.75 \cdot 10^{\lfloor \log_{10}( n\_agents_k ) \rfloor} \rceil \\
    n\_cities_k & := \left\lfloor \frac{n\_agents_k}{10} \right\rfloor + 2 \\
    x\_dim_k & := \left\lceil \sqrt{6 \left( \left\lceil \frac{max\_rails\_in\_city}{2} \right\rceil + 3 \right)^2 \cdot n\_cities_k} \right\rceil + 7 \\
    y\_dim_k & := x\_dim_k
\end{align*}

In total, 41 test $k\in\{0, \ldots, 40\}$ were generated, each containing 10 environments $l\in\{0, \ldots, 9\}$, which shared the same dimensions, number of agents $n_{k}\in[1,6256]\cap\mathbb{N}$, and other parameters (see above).
Dimensions ranged from $25\times25$ to $314\times314$, all square.
The environments in each level had malfunction rates $r = ( l \cdot min\_malfunction\_interval )^{-1}, $ for $l>0$, and $r=0$ for $l=0$.

\section{Evaluation Setup}
\label{app:eval_setup}
The submissions were evaluated in a docker container with access to 4 CPU cores and 15 GB of main memory. Each core was an hyper-thread of an Intel Xeon E5 v3 with a base speed of 2.3 GHz and a single-core maximum turbo speed of 3.8 GHz. The base image for the container was Ubuntu 18.04.

The submissions were built in advance of the evaluation when necessary, for example to compile C/C++ code. They could only communicate with the evaluation environment through a restricted interface to avoid any abuse.

\section{Graph Representation of the Flatland Environment}
\label{app:graph}

\begin{figure}[h]
    \centering
    \includegraphics[width=1.0\linewidth]{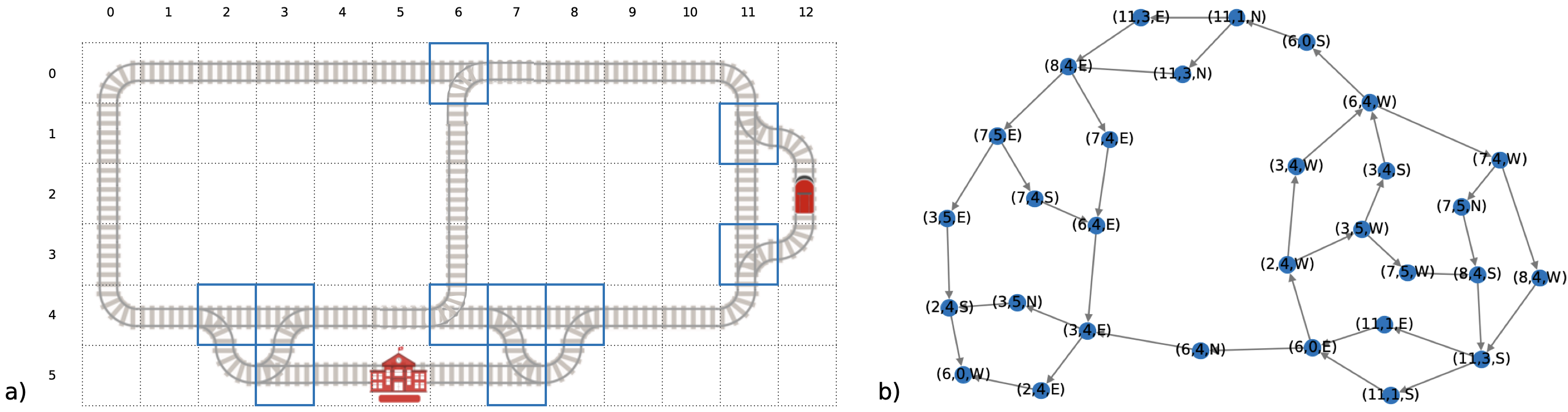}
    \caption{Visualization of a graph representation of the environment where a vertex represents a switch and the direction of entry.}
    \label{fig:graph}
\end{figure}

A Flatland environment can be represented as different kinds of graphs.
For example, every cell can be modelled, but also graphs that only include decision cells are useful.
For most applications, not only the rail layout but also the available direction options an agent has at a vertex, dependent on its incoming direction (edge), have to be modelled.
In that case, a cell is translated into multiple vertices, one for each direction available.
\newpage
\section{Outstanding Solutions Hyper-parameters}
\label{app:solution_configs}


\subsection{Team JBR\_HSE}

\subsubsection{Reinforcement Learning Model}

The model was trained using PPO \citep{PPO}.

\medskip

\begin{tabular}{|l|l|}
\hline
Maximum training timesteps & $10^{10}$ \\ \hline
Maximum training episodes & $10^{5}$   \\ \hline
Optimizer              & Adam                   \\ \hline
Learning rate          & $10^{-5}$ \\ \hline
Batch size             & 32                     \\ \hline
GAE horizon            & 16                     \\ \hline
GAE gamma              & 0.995                  \\ \hline
GAE lambda             & 0.95                   \\ \hline
Epsilon (clipping)     & 0.2                    \\ \hline
Value loss coefficient & 0.5                    \\ \hline
Entropy coefficient    & $10^{-2}$              \\ \hline
Actor network    & [256, 128] fully connected              \\ \hline
Critic network    & [256, 128] fully connected              \\ \hline
Network activations    & ReLU              \\ \hline
Model architecture    & See \sectionref{sec:rl1}              \\ \hline
\end{tabular}

\subsubsection{Departure Classifier}

The classifier limited the maximum occupancy by only allowing trains that have a high probability to reach their target to enter the grid.

\medskip

\begin{tabular}{|l|l|}
\hline
Training epochs        & 3 \\ \hline
Optimizer              & Adam                   \\ \hline
Learning rate          & $10^{-4}$ \\ \hline
Batch size             & 8                     \\ \hline
Probability threshold  & 0.92                     \\ \hline
Network layers    & [128, 128] fully connected              \\ \hline
Network activations    & ReLU              \\ \hline
\end{tabular}

\newpage
\subsection{Team Netcetera}

The model was trained using Ape-X DQN with Duelling Networks \citep{horgan2018distributed}.

\medskip

\begin{tabular}{|l|l|}
\hline
Training timesteps        & 2,000 \\ \hline
Optimizer              & Adam                   \\ \hline
Learning rate          & 5e-4 \\ \hline
Batch size             & 128                     \\ \hline
Epsilon (exploration)  & 1 to 0.02 annealed over 15,000 timesteps       \\ \hline
Network layers    & [30, 30, 30, 30, 30, 20, 20, 20, 20, 10, 5] fully connected             \\ \hline
Network activations    & ReLU              \\ \hline
\end{tabular}

\medskip

\noindent The other parameters used the values provided by the RLlib Ape-X DQN implementation.\footnote{\url{https://docs.ray.io/en/releases-0.8.7/rllib-algorithms.html}}

\subsection{Team MARMot-Lab-NUS}

The model was trained using A3C~\citep{mnih2016asynchronous}.

\medskip

\begin{tabular}{|l|l|}
\hline
Optimizer              & Nadam                   \\ \hline
Learning rate          & $2^{-5}$ \\ \hline
Gamma             & .95                     \\ \hline
Network layers    & [256, 256, 256, 256, 128, 64] convolution, [1024] fully connected              \\ \hline
Network activations    & ReLU              \\ \hline
\end{tabular}

\section{Cell Types in a Flatland Environment}
\label{app:cell_types}

\begin{figure}[!h]
    \centering
    \includegraphics[width=0.6\linewidth]{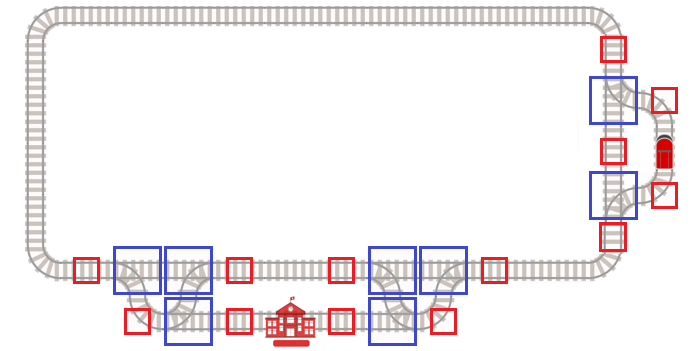}
    \caption{\textbf{Cell types.} There are three different types of cells in a Flatland environment (see \sectionref{sec:masking}). The majority of cells containing rails in a Flatland environment are non-decision cells (grey rails). The blue cells are decision cells where an agent has multiple options (depending on the incoming direction) in which direction to proceed. Directly connected decision cells form clusters which play an important role for train coordination. Red cells denote stopping cells that are connected to decision cells but not decision cells themselves.}
    \label{fig:cell_types}
\end{figure}
\newpage

\section{JBR\_HSE Solution Architecture}
\label{app:architecture}

\begin{figure}[!h]
    \centering
    \includegraphics[width=0.65\linewidth]{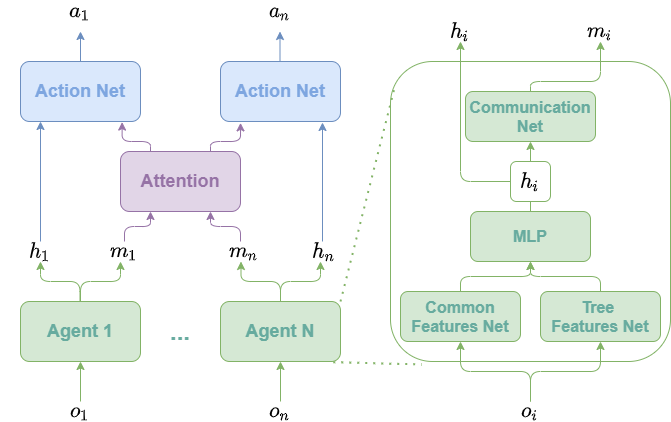}
    \caption{\textbf{Agent architecture.} $o, a, m$, and $h$ respectively denote observations, actions, messages, and extracted features.}
    \label{fig:architecture}
\end{figure}

\section{Environment Completion Rates}
\label{app:completation}

\begin{figure}[!h]
    \centering
    \includegraphics[width=0.9\linewidth]{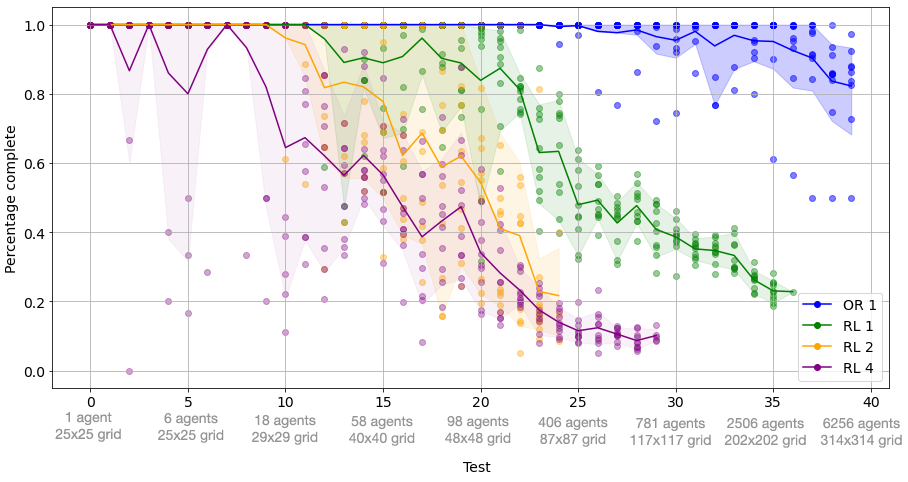}
    \caption{\textbf{Ratio of agents that reached their targets per environment.} The dots indicate the rate of agents that completed a given environment. The lines represent the mean of the 10 completion rates per test (x-axis) and the filled area the 0.1-quantile.}
    \label{fig:completation}
\end{figure}
\newpage
\section{Visual Analytics for Flatland}
\label{app:visualization}

\begin{figure}[ht]
    \centering
    \includegraphics[width=0.75\linewidth]{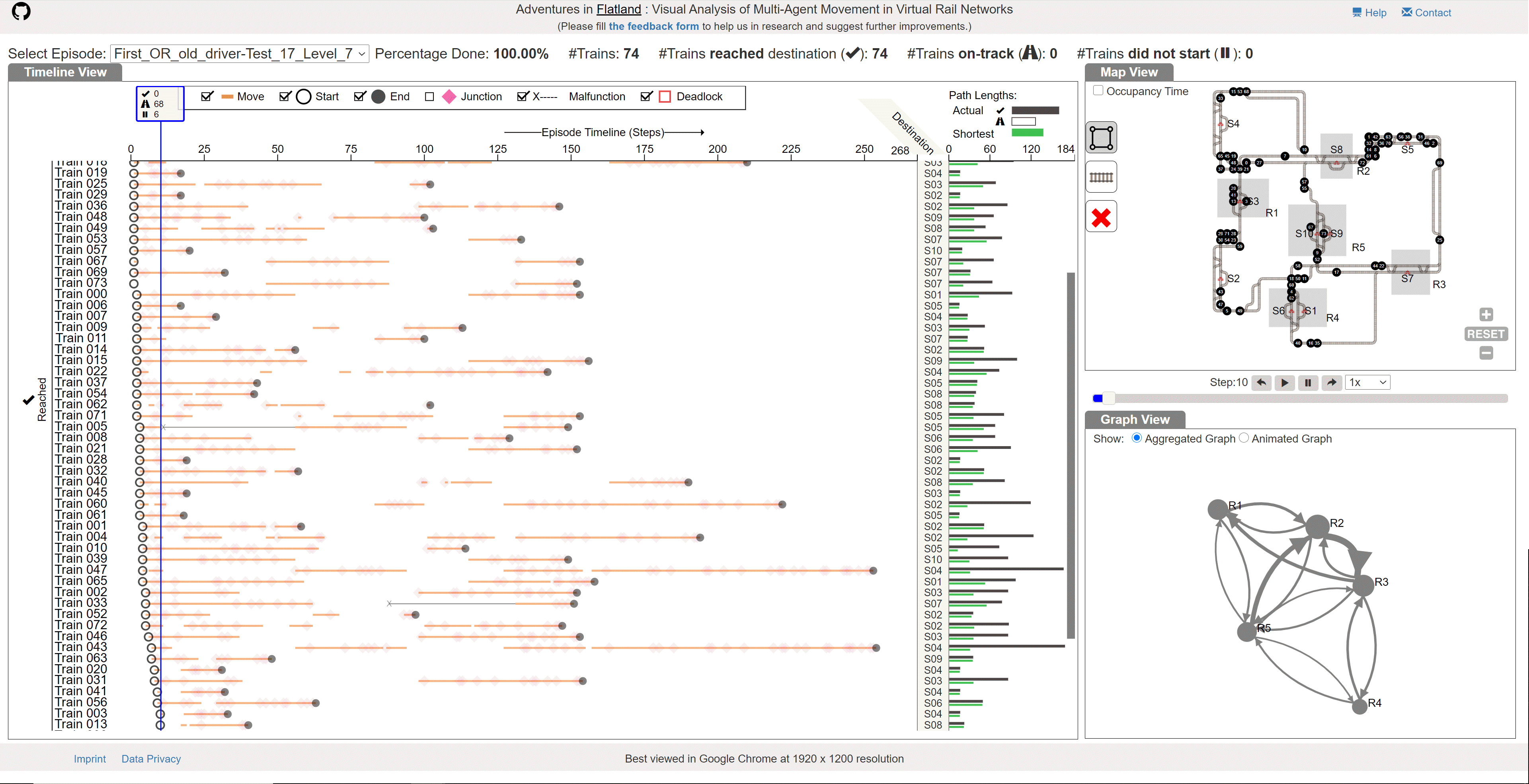}
    \caption{\textbf{A visual analytics approach to analyze agent behaviour.} The timeline view is shown on the left, the map view on the top right, and the graph view on the bottom right.}
    \label{fig:visualization}
\end{figure}

\begin{figure}[ht]
    \centering
    \includegraphics[width=0.75\linewidth]{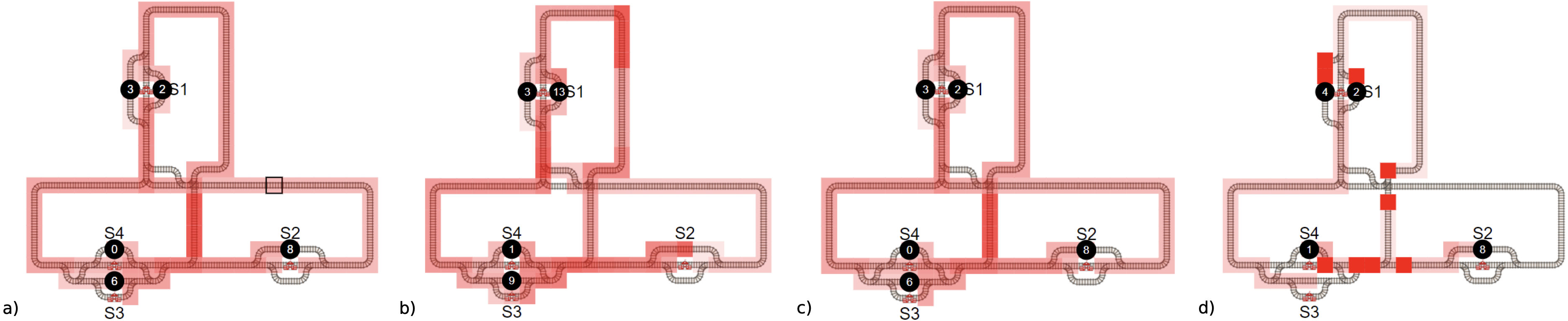}
    \caption{\textbf{Visual analysis of the track usage.} In this example (environment 7 of test 10), An\_Old\_Driver (a), JBR\_HSE (b), and Netcetera (c) showed similar behavior by utilizing almost all the tracks (red heatmap), whereas MARMot-Lab-NUS (d) used fewer tracks. JBR\_HSE and MARMot-Lab-NUS both experienced deadlocks (red solid blocks). Individual trains are represented through numbered black circles and stations are labelled S1-S4.}
    \label{fig:tracks}
\end{figure}

The visual analytics tool\footnote{\url{https://github.com/shivamworking/flatland-visualization}} offered three static and interlinked views to investigate the performance of a solution in an individual environment (see \figureref{fig:visualization}): the timeline view, the map view and the graph view. The timeline view allowed to investigate actions and events (e.g., movement, malfunctions, deadlock, etc.) for each train in a separate row. The map view highlighted the train positions on tracks for individual timesteps and the graph view showed the user-defined regions-of-interest from the map view as vertices and agent movements between the regions as edges. \figureref{fig:tracks} shows an example of a visual analysis of the track usage.

\end{document}